\documentclass[sigconf]{acmart}
\AtBeginDocument{%
  }

\usepackage{latexsym}
\usepackage{microtype}
\usepackage{multirow}
\usepackage{xcolor}
\usepackage[most]{tcolorbox}
\usepackage{bbm}
\usepackage[bottom]{footmisc}
\usepackage{float}
\usepackage{tikz}
\usepackage{xurl}
\usepackage{pgfplots}
\usepgfplotslibrary{groupplots}
\usepackage{graphics}
\usepackage{longtable}
\usepackage{caption}
\usepackage{hyperref}
\usepackage{subcaption}
\usepackage{algorithm}
\usepackage{algpseudocode}
\usepackage{booktabs}
\usepackage{graphicx}
\usepackage{amsmath}
\usepackage{balance}
\usepackage{enumitem}
\usepackage{pifont}
\pgfplotsset{compat=1.18}
\definecolor{CustomBlue}{RGB}{20, 81, 124}  
\definecolor{CustomLightBlue}{RGB}{47, 127, 193}  
\definecolor{CustomVeryLightBlue}{RGB}{231, 239, 250}  
\definecolor{CustomGreen}{RGB}{150, 195, 125}  
\definecolor{CustomYellow}{RGB}{243, 210, 102}  
\definecolor{CustomRed}{RGB}{216, 56, 58}  
\definecolor{CustomPink}{RGB}{247, 225, 237}  
\definecolor{CustomVeryLightPurple}{RGB}{248, 243, 249}  
\definecolor{CustomPurple}{RGB}{196, 151, 178}  
\definecolor{CustomGrayBlue}{RGB}{169, 184, 198}  
\begin{document}

\title{ELOQ: Resources for Enhancing LLM Detection of Out-of-Scope Questions}


\author{Zhiyuan Peng}
\authornote{Both authors contributed equally to this research.}
\affiliation{%
  \institution{Santa Clara University}
  \city{Santa Clara}
  \state{CA}
  \country{USA}}
\email{zpeng@scu.edu}

\author{Jinming Nian}
\authornotemark[1]
\affiliation{%
  \institution{Santa Clara University}
  \city{Santa Clara}
  \state{CA}
  \country{USA}}
\email{jnian@scu.edu}

\author{Alexandre Evfimievski}
\authornote{The author contributed to this work before he joined Adobe Inc.}
\affiliation{%
  \institution{Adobe Inc.}
  \city{San Jose}
  \state{CA}
  \country{USA}}
\email{aevfim@outlook.com}

\author{Yi Fang}
\affiliation{%
  \institution{Santa Clara University}
  \city{Santa Clara}
  \state{CA}
  \country{USA}}
\email{yfang@scu.edu}

\renewcommand{\shortauthors}{Peng et al.}

\begin{abstract}

Retrieval-augmented generation (RAG) has become integral to large language models (LLMs), particularly for conversational AI systems where user questions may reference knowledge beyond the LLMs’ training cutoff. However, many natural user questions lack well-defined answers, either due to limited domain knowledge or because the retrieval system returns documents that are relevant in appearance but uninformative in content. In such cases, LLMs often produce hallucinated answers without flagging them. While recent work has largely focused on questions with false premises, we study out-of-scope questions, where the retrieved document appears semantically similar to the question but lacks the necessary information to answer it. In this paper, we propose a guided hallucination-based approach ELOQ \footnote{\url{https://github.com/zhiyuanpeng/ELOQ.git}}, to automatically generate a diverse set of out-of-scope questions from post-cutoff documents, followed by human verification to ensure quality. We use this dataset to evaluate several LLMs on their ability to detect out-of-scope questions and generate appropriate responses. Finally, we introduce an improved detection method that enhances the reliability of LLM-based question-answering systems in handling out-of-scope questions. 
\end{abstract}

\ccsdesc[500]{Computing methodologies~Natural language processing}
\ccsdesc[500]{Information systems~Question answering}

\keywords{Large Language Models, Question Answering, Out-of-Scope Question, Retrieval Augmented Generation}


\maketitle

\section{Introduction}
Retrieval Augmented Generation (RAG) has become a standard approach to building context-grounded conversational AI agents~\cite{Lewis2020:RAG, Guu2020:RAG, Gao2024:RAGSurvey}. Given an inquiry, a RAG system retrieves a relevant document from a curated knowledge base and presents it to a large language model (LLM), which generates a response. RAG performance is evaluated on dimensions such as response accuracy, faithfulness, completeness, answer and context relevance, and style \cite{Es2024:RAGAS, SaadFalcon2024:ARES, Liu2024:calibrating, Zheng2023:judging}. Errors may stem from a document-to-question mismatch, hallucinated parts of the response, or misunderstanding of the document or the question.

Many user questions have no direct answer: some rely on a false premise, others are ambiguous in their intent, and some are out-of-scope for the knowledge base. Prior studies indicate that about 25\% of natural questions contain false assumptions~\cite{Yu2023:CREPE} and over 50\% are ambiguous~\cite{DBLP:conf/emnlp/MinMHZ20}. Additionally, out-of-scope errors have been identified as the leading cause of enterprise AI assistants generating responses that appear convincing but are, in fact, incorrect~\cite{Maharaj2024:evaluation}. A well-designed RAG system should implement an exception handling mechanism that detects confusing questions and responds by clarifying the confusion, retrieving additional documents, or seeking assistance. 

\begin{table}[t]
\centering
\resizebox{0.5\textwidth}{!}{  
\begin{tabular}{c|cccc}
\toprule
Dataset & Source & Category & Cutoff & Annotation Method \\
\midrule
SQuAD 2.0~\cite{DBLP:conf/acl/RajpurkarJL18} & Wiki      & Unanswerable   & \textcolor{CustomRed}{\ding{55}} & Human \\
SQuAD2-CR~\cite{DBLP:conf/lrec/LeeHC20} & Wiki      & Unanswerable   & \textcolor{CustomRed}{\ding{55}} & Human \\
BoolQ${}_{3L}$~\cite{DBLP:conf/naacl/SulemHR22} & Wiki      & Unanswerable   & \textcolor{CustomRed}{\ding{55}} & Algoritm+Human \\
CAsT-answerability~\cite{DBLP:conf/ecir/LajewskaB24} & Wiki      & Unanswerable   & \textcolor{CustomRed}{\ding{55}} & Algoritm+Human \\
AMBIGQA~\cite{DBLP:conf/emnlp/MinMHZ20} & Wiki      & Multi-answers   & \textcolor{CustomRed}{\ding{55}} & Human \\
CREPE~\cite{Yu2023:CREPE} & Reddit & False Premise & \textcolor{CustomRed}{\ding{55}} & Human \\
FalseQA~\cite{Hu2023:fooled}  & --- &\ False Premise & \textcolor{CustomRed}{\ding{55}} & Human \\
KG-FPQ~\cite{Zhu2024:KGFPQ}     & Wiki      & False Premise   & \textcolor{CustomRed}{\ding{55}} & LLM \\
FAITH~\cite{Yuan2024:whispers}  & Wiki      & False Premise   & \textcolor{CustomRed}{\ding{55}} & LLM \\
FLUB~\cite{li2024llms}          & Baidu Tieba     & Cunning Texts         & \textcolor{CustomRed}{\ding{55}} & Human \\

ELOQ                         & News      & Out-of-scope    & \textcolor{CustomGreen}{\ding{52}} & LLM+Human \\
\bottomrule
\end{tabular}
}
\caption{ELOQ vs Existing Datasets. ``Cutoff'' represents whether the data is beyond the LLMs' knowledge or not.}
\label{tab: dataset_comparison}
\end{table}

Enabling “deconfusion” in a RAG system involves three challenges: (1) creating a diverse dataset of confusing questions, (2) training a classifier to detect confusion, and (3) developing a response generator for each confusion type. 
As shown in Table~\ref{tab: dataset_comparison}, prior studies create confusing questions with different categories, such as multi-answer~\cite{DBLP:conf/emnlp/MinMHZ20}, false premise~\cite{Yu2023:CREPE, Hu2023:fooled, Zhu2024:KGFPQ, Yuan2024:whispers}, and cunning texts~\cite{li2024llms}. For instance, ``How many sacks does clay matthews have in his career'' is a multi-answer question as ``clay matthews'' may refer to ``Clay Matthews Jr.'' or ``Clay Matthews III'', ``Where in Liverpool did John Lennon die'' is a false premise question as John Lennon died in New York city and ``What should I do if I forget which ATM machine I deposited my money in?'' is a cunning text. In contrast to these questions, out-of-scope questions look relevant to the document, but there is no answer within the document and thus can mislead LLMs into generating hallucinated answers that seem correct but are actually incorrect, representing the most dangerous type of error~\cite{Maharaj2024:evaluation}. For instance, ``How does Justin Jefferson’s exceptional catching ability influence the Vikings’ decision to select him in the first round of the 2020 NFL Draft despite his lack of speed?'' is an out-of-scope question as it asks about ``Vikings’ decision to select Jefferson in the first round of the 2020 NFL Draft'' which was never discussed in the paired news (Appendix~\ref{ap: exp_oos_doc}). Our work focuses on mitigating this type of confusion. 

Existing studies~\cite{DBLP:conf/acl/RajpurkarJL18, DBLP:conf/lrec/LeeHC20, DBLP:conf/naacl/SulemHR22, DBLP:conf/ecir/LajewskaB24} on unanswerable questions of which out-of-scope questions are a subset, have typically created the dataset either through manual annotation by trained human annotators~\cite{DBLP:conf/lrec/LeeHC20, Hu2023:fooled} or via simple automatic generation algorithms, as seen in BoolQ${}_{3L}$~\cite{DBLP:conf/naacl/SulemHR22} and CAsT-answerability~\cite{DBLP:conf/ecir/LajewskaB24}. These methods often lack a balance between efficiency and quality, as human-annotated questions are of high quality but low efficiency, and automatically generated questions are of high efficiency but too simple to 
be distinguished to some degree. Recently, LLMs have demonstrated strong generation and following instruction abilities, and thus KG-FPQ~\cite{Zhu2024:KGFPQ} and FAITH~\cite{Yuan2024:whispers} utilize LLMs to generate high-quality false premise questions, mitigating the balance of efficiency and quality issues. However, all these datasets are usually constrained to knowledge before the LLMs’ training cutoff date, which limits their effectiveness in evaluating real-world RAG systems that must retrieve up-to-date information likely beyond the model's training data to answer user questions. 

To address these challenges above, we propose a framework for automatically generating out-of-scope questions and evaluate its effectiveness in training classifiers to detect out-of-scope situations. While questions in existing datasets such as MS MARCO \cite{DBLP:conf/nips/NguyenRSGTMD16}, may serve as out-of-scope questions for the corresponding hard negative passages, they fail beyond the LLM’s knowledge cutoff. ELOQ fills this gap by providing a diverse benchmark for out-of-scope questions, leveraging LLM-assisted generation to reduce human annotation effort. Our contributions are as follows:

\begin{itemize}[topsep=4pt,itemsep=1pt,parsep=2pt,leftmargin=9pt]
\item We propose a framework for generating synthetic out-of-scope questions from a given corpus.
\item We demonstrate the utility of our synthetic data in training out-of-scope question detectors.
\item We conduct a comparative evaluation of multiple LLMs as RAG agents to measure their out-of-scope question detection accuracy.
\end{itemize}

\section{Related Work}

\subsection{Benchmark Datasets}
Recent work on LLM responses to confusing questions primarily focuses on stand-alone questions answered using the LLM's pretraining knowledge, without a supporting context document. These questions may be naturally collected from online sources~\cite{Li2024:cunning,Yu2023:CREPE}, written by human annotators~\cite{Hu2023:fooled}, or synthetically generated~\cite{Zhu2024:KGFPQ,Yuan2024:whispers}. 
Earlier research on ambiguous open-domain questions introduced AmbigQA~\cite{Min2020:AmbigQA}, a dataset of naturally ambiguous questions. More relevant benchmarks are unanswerable questions~\cite{DBLP:conf/acl/RajpurkarJL18, DBLP:conf/lrec/LeeHC20, DBLP:conf/naacl/SulemHR22, DBLP:conf/ecir/LajewskaB24} of which out-of-scope questions are a subset. SQuAD 2.0~\cite{DBLP:conf/acl/RajpurkarJL18} is created by asking crowdworkers to write unanswerable questions similar to the answerable question given a paragraph. SQuAD2-CR~\cite{DBLP:conf/lrec/LeeHC20} tags SQuAD 2.0~\cite{DBLP:conf/acl/RajpurkarJL18} with answerable reasons. CAsT-answerability~\cite{DBLP:conf/ecir/LajewskaB24} extends CAsT-snippets~\cite{Lajewska:2023:CIKM} by including five randomly selected non-relevant passages to each question. BoolQ${}_{3L}$~\cite{DBLP:conf/naacl/SulemHR22} matches each selected ``Yes'' or ``No'' question of BoolQ~\cite{DBLP:conf/naacl/ClarkLCK0T19} to a passage within BoolQ that has the greatest overlap with the questions in terms of nouns and verbs. These unanswerable datasets are sourced from Wikipedia, which is typically included in the training data of LLMs. Our dataset is crawled from news, allowing us to easily select news published only after the knowledge cutoff date of LLMs. Table~\ref{tab: dataset_comparison} presents a comparison of these benchmark datasets.

\subsection{Question Generation}
\citet{Zhu2024:KGFPQ} and \citet{Yuan2024:whispers} describe a method to synthetically generate stand-alone false premise questions by selecting a set of factual triples of (subject, relation, object) using Wikipedia data, replacing the object with a similar but incorrect one, and then filling in a template to generate a false premise question. Content-grounded (non-confusing) synthetic data generation has been studied in \cite{Yehudai2024:parity,Wang2023:selfinstruct,Lee2023:ensemble}. Given the grounding passages, these methods generate data entries using few-shot prompting with task-specific examples, followed by filtering to ensure data quality, faithfulness, and diversity.

\subsection{Mitigation}
Mitigation is commonly achieved through in-context learning ~\cite{Zhu2024:KGFPQ,Li2024:cunning}, chain-of-thought (CoT) ~\cite{DBLP:conf/nips/Wei0SBIXCLZ22} and fine-tuning. Experiments show that fine-tuning outperforms both in-context learning and CoT reasoning~\cite{Hu2023:fooled}. CoT reasoning leads to inconsistent improvements, whereas in-context learning improves performance as the number of examples increases~\cite{Li2024:cunning}. 
\citet{Yuan2024:whispers} identifies approximately 1\% of attention heads as responsible for confusion caused by false-premise questions and demonstrates a 20\% performance gain by constraining these heads.
~\citet{Kim2021:lightbulb} retrieves a Wikipedia page for each question and responds with the identified unverifiable premises if the page does not entail them. ~\citet{DBLP:conf/naacl/SulemHR22} trains a classification model based on BERT to identify unanswerable questions.

\begin{algorithm}[t!]
\small
\caption{The guided hallucination method}\label{alg:ghal}
\begin{algorithmic}[1]
\State Call LLM${}_q$ to \hyperref[ap: extract_prompt]{extract} from~$d$ a list of $n$ claims $c_1, \ldots, c_n$; batching claims reduces LLM${}_q$-calls (Appendix~\ref{ap: extract_prompt})
\State Partition $\{c_1, \ldots, c_n\}$ into disjoint subsets $S_1, \ldots, S_k$, e.g.\ $S_j = \{c_i: i \mathop{\textrm{mod}} 3 = j - 1\}$ for $j = 1, 2, 3$
\For{several rounds (e.g.\ 3), and in each round, for all $S_j$ in partition $(S_1, \ldots, S_k)$}
    \State In the claims list $c_1, \ldots, c_n$ replace all claims $c_i$ where $i\in S_j$ by text ``\texttt{(missing)}''
    \State Provide LLM${}_q$ with this modified list (but not~$d$) and prompt it to \hyperref[ap: recover_prompt]{recover} the missing claims (Appendix~\ref{ap: recover_prompt})
    \State In the claims list $c_1, \ldots, c_n$, replace the missing claims with the claims recovered in the above step.
\EndFor
\State Call LLM${}_q$ to \hyperref[ap: prompt_remove_claim]{remove} all claims supported by~$d$ or by the original claims, leaving only out-of-scope claims (Appendix~\ref{ap: prompt_remove_claim})
\State \label{ghal:genq}Call LLM${}_q$ to \hyperref[ap: prompt_conf_q_gen]{generate} one short question per each novel claim, focusing on one key element of it (Appendix~\ref{ap: prompt_conf_q_gen})
\State Call LLM${}_q$ to \hyperref[ap: prompt_confusion]{filter} out any questions answerable in the context of~$d$ (Appendix~\ref{ap: prompt_confusion})
\end{algorithmic}
\end{algorithm}

\section{Methods}

We developed the ELOQ data generator to evaluate and mitigate LLM confusion when responding to out-of-scope questions for a given document. We assume that the user has limited domain knowledge, resulting in asking questions that appear relevant but are actually out of scope and cannot be answered based on the document’s content. We also assume that, due to the limitations of the retriever itself, relevant documents without an answer to the questions are possibly retrieved and ranked at the top. The best answer to this kind of question is to refuse to answer it, rather than providing a fabricated response. In this step, given a standard prompt $p$, an out-of-scope question~$q$, and a relevant document~$d$, the LLM${}_r$ generates a response~$r$. Ideally, $r$ should clarify the confusing part of the question -- a process we refer to as defusion (or de-confusion) -- rather than attempting to answer the question directly and risking hallucination.

\subsection{Data Collection}\label{sec:data_collection}

We focus on the scenario where the domain knowledge is maintained separately from the LLM in a document database. Hence, we prefer documents that are novel to the evaluated LLMs, such as news articles published after all pretraining cutoff dates, ensuring that LLMs cannot reproduce claims or answer questions correctly by pure hallucination. Instead, they must generate responses grounded in their understanding of the news content. Therefore, using the Newscatcher~\cite{WNewscatcher}, we collected 200 news articles published between Jan 1, 2024, and September 30, 2024, for each topic\footnote{sport, business, science, food, politics, travel, entertainment, music, news, tech}. We require each document to be concise enough to fit within the LLM prompt, along with instructions and examples, but sufficiently long to make at least 4 to 10 separate claims (i.e., a few paragraphs in length). Thus, we select news articles with more than 150 words and, for each, sequentially extract sentences from the beginning until the word count exceeds 300. Finally, we collected 2,000 news and sampled a subset for human annotation. We represent the human-annotated data as Gold and the remaining data as Silver. Table~\ref{tab:statistic} presents the statistics of the dataset. Our ELOQ dataset \footnote{\url{https://huggingface.co/datasets/zhiyuanpeng/ELOQ}} is publicly available.

\subsection{Question Generation} \label{sec: data_gen}
We introduce our guided hallucination method, designed to generate out-of-scope questions from a given document. The process consists of three main steps: claim extraction, hallucination injection, and question generation. 

\subsubsection{Claim Extraction}\label{sec:claims_extraction}
The first step involves extracting claims from the document $d$ to serve as the basis for generating out-of-scope questions. Since our goal is to generate questions that appear related to $d$ but are actually unanswerable using its content, we start by obtaining a structured set of claims from the document. To achieve this, we use a LLM denoted as  LLM${}_q$, to extract a set of $n$ claims $c_1, c_2, …, c_n$ from $d$. These claims represent key factual statements or assertions made in the document. We extract the claims in batches to minimize the number of LLM calls and improve computational efficiency. Once the claims are extracted, we partition them into disjoint subsets $S_1, S_2, …, S_k$. A simple partitioning scheme is used, such as:
$$
S_j = \{c_i \mid i \mod 3 = j - 1\}, \quad \text{for } j = 1, 2, 3.
$$
This ensures that claims are distributed evenly among the subsets, facilitating an efficient iterative process in the next step. Partitioning the claims allows us to selectively manipulate certain claims while preserving the overall document structure.

\subsubsection{Hallucination Injection}
The second step is to generate hallucinated claims that are similar to the factual claims generated in Section~\ref{sec:claims_extraction} but contain altered or missing information. The process follows an iterative masking and recovery approach over multiple rounds (e.g. 3). For each round, we follow steps 3 to 6 in Algorithm~\ref{alg:ghal}.

In practice, we find that even after repeating the hallucination injection process multiple times, some claims remain too general and can still be supported by the document. Therefore, we design a prompt-based filter ~\ref{ap: prompt_remove_claim} to discard these in-scope claims that are explicitly supported by $d$. This ensures that only truly out-of-scope claims remain in our dataset.

\begin{table}[t]  
\centering
\begin{tabular}{l|cc}
\toprule
Attribute & Gold & Silver\\
\midrule
 \# of documents & 45 & 1,955 \\
 \# of in-scope questions & 103 & 8,949 \\
 \# of out-of-scope questions & 113 & 10,105 \\
 average words per document & 280 & 276 \\
 average words per in-scope question & 16 & 16 \\
 average words per out-of-scope question & 16 & 16 \\
\bottomrule
\end{tabular}
\caption{Statistics of ELOQ. Gold is sampled from crawled 2,000 news and annotated by annotators (Section~\ref{sec: human_annotation}).}
\label{tab:statistic}
\end{table}

\subsubsection{Question Generation} 
The final step converts the hallucinated claims into out-of-scope questions while ensuring they cannot be answered using the document. Specifically, for each hallucinated claim, LLM${}_q$ generates a concise and precise question focusing on a key aspect of the claim. The question is designed to appear related to the document while being unanswerable based on its content. Once the questions are generated, we further filter them by ensuring they are truly unanswerable. This is achieved by prompting LLM${}_q$ to verify whether the generated questions can be answered using $d$. Any question that remains answerable in the context of $d$ is discarded. 

In addition to generating out-of-scope questions, we also prompt LLM${}_q$ to generate in-scope questions for comparison. The complete question generation process is outlined in Algorithm~\ref{alg:ghal}, and the associated prompts are in Appendix~\ref{ap: prompt}.

\begin{table*}[t!]
\label{tab:human_annotation}
\centering
\resizebox{\textwidth}{!}{  
\begin{tabular}{c|c|cc cc cc cc}
\toprule
\multirow{2}{*}{\textbf{Group}} & \multirow{2}{*}{\textbf{Annotator}} & \multicolumn{2}{c}{\textbf{Cohen's Kappa}} & \multicolumn{2}{c}{\textbf{Annotator Acc}} & \multicolumn{2}{c}{\textbf{Group's Agree Acc}} & \multicolumn{2}{c}{\textbf{Ground Truth Acc}} \\ 
\cmidrule(lr){3-4} \cmidrule(lr){5-6} \cmidrule(lr){7-8} \cmidrule(lr){9-10}  
&& \textbf{Out-of-Scope}& \textbf{Defusion}&\textbf{Out-of-Scope}& \textbf{Defusion}&\textbf{Out-of-Scope}& \textbf{Defusion}& \textbf{Out-of-Scope}& \textbf{Defusion}\\ 
\midrule
\multirow{2}{*}{1} & A1 & \multirow{2}{*}{0.8053} & \multirow{2}{*}{0.8917} & 88.89 & 97.37 & \multirow{2}{*}{92.31} & \multirow{2}{*}{96.97} & \multirow{6}{*}{94.91} & \multirow{6}{*}{98.23} \\ 
& A2 &  & & 87.50& 94.59& & &&\\ 
\cline{1-8}
\multirow{2}{*}{2}& A3 & \multirow{2}{*}{0.7508} & \multirow{2}{*}{0.8187} & 93.06& 97.14& \multirow{2}{*}{98.41}& \multirow{2}{*}{96.77}&  &  \\ 
& A4 &  &  & 91.67& 90.00& & &&\\ 
\cline{1-8}
\multirow{2}{*}{3}& A5 & \multirow{2}{*}{0.8615} & \multirow{2}{*}{0.8333} & 91.67& 93.75& \multirow{2}{*}{98.51}& \multirow{2}{*}{100.00}&  &  \\ 
& A6 &  &  & 98.61& 100.00& & &&\\ 
\bottomrule
\end{tabular}
}
\caption{Agreement scores and the performance of our method on the ELOQ-Gold. ``Annotator'' represents the annotator's name. ``Out-of-Scope'' represents whether the question is out-of-scope or not. ``Defusion'' refers to whether an LLM's response indicates that the question has no suitable answer based on the document.}
\label{tab: human_label}
\end{table*}

\begin{algorithm}[t]
\caption{Evaluation steps}\label{alg:eval}
\begin{algorithmic}[1]
\State \label{eval:rag}Call LLM${}_r$ on $(d, q)$ with RAG prompt and get its response~$r$ (Appendix~\ref{ap: prompt_rag}, ~\ref{ap: prompt_rag_twoshot}, ~\ref{ap: prompt_rag_cot})
\State \label{eval:cnf}Call LLM${}_r$ $m$~times on $(d, q)$ to check and explain if~$q$ is out-of-scope given~$d$ (Appendix~\ref{ap: prompt_confusion})
\State Aggregate $m$ predictions by majority vote
\If{$q$ is generated as out-of-scope}
  \State \label{eval:def}Call LLM${}_q$ $m$~times on $(d, q, r)$ to check and explain if~$r$ defused the confusion (Appendix~\ref{ap: prompt_defusion})
  \State Aggregate $m$ predictions by majority vote
\EndIf
\end{algorithmic}
\end{algorithm}

\subsection{Evaluation} \label{sec: eval_metrics}
We first assessed the quality of ELOQ by sampling a subset of questions (ELOQ-Gold) and manually verifying whether they were truly out-of-scope (Section \ref{sec:humann}). Next, we evaluated how well various LLM${}_r$ models could defuse the out-of-scope questions in ELOQ. Given the large number of out-of-scope questions, manual verification of LLM${}_r$ responses is not feasible. Therefore, we proposed AutoDefuseEval using GPT-4o-mini to automatically detect whether a response successfully defuses the question. To validate AutoDefuseEval, we compared its performance with human annotations (ELOQ-Gold) on a sampled dataset (Section \ref{sec:humann}). Formally, we define our tasks as follows:
\begin{description}[topsep=2pt,itemsep=0pt,parsep=2pt,leftmargin=0pt]
\item[Question generation:] What is the quality of out-of-scope and in-scope questions generated by Algorithm~\ref{alg:ghal}? (Section \ref{sec:humann})
\item[Defusion detection:] How accurately can AutoDefuseEval detect whether an LLM${}_r$ response to a out-of-scope question defuses the confusion? (Section \ref{sec:humann})
\item[Out-of-scope response:] How often does an LLM${}_r$ successfully defuse an out-of-scope question using different prompting methods? (Section \ref{sec: exp_conf_response})
\item[Out-of-scope detection:] How accurately can LLMs identify which context-grounded questions are out-of-scope and require special handling? (Section \ref{sec: exp_conf_detection})
\end{description}

Algorithm~\ref{alg:eval} has the steps we run for evaluating the above tasks. We run steps 1, 2, and 3 for ``out-of-scope detection'' and 1, 4, 5, 6, and 7 for ``out-of-scope response''. 
Following the self-consistency method~\cite{Wang2023:selfconsistency}, we perform multiple LLM calls in steps ~\ref{eval:cnf} and ~\ref{eval:def} of Algorithm~\ref{alg:eval}, taking the majority vote to determine the final label. 

\subsection{Human Annotation}\label{sec: human_annotation}
\label{sec:humann}
\begin{figure} [h] 
  \label{fig:human}
  \centering
  \includegraphics[width=0.9\linewidth]{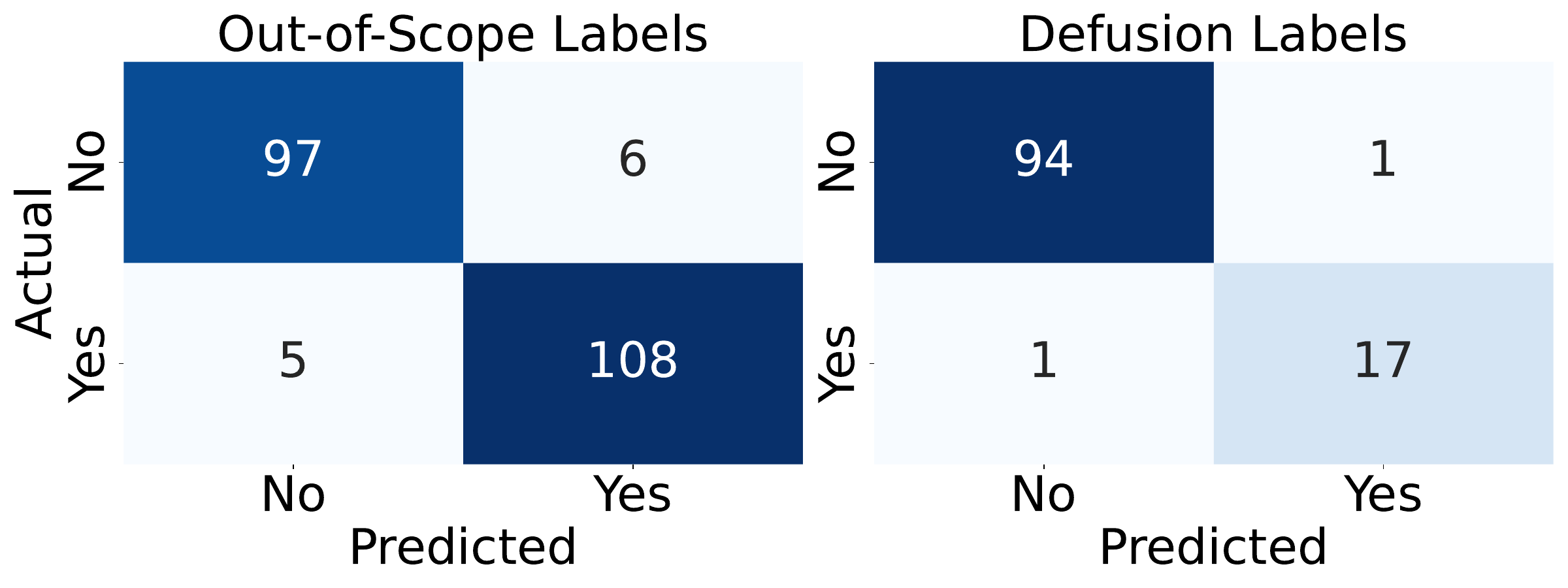}
  \caption{Confusion matrix of out-of-scope and defusion on ELOQ-Gold.}
  \label{fig:confuse_defuse}
\end{figure}

Following the methodology from \cite{DBLP:books/wi/Cochran77}, we sampled 216 questions, evenly divided into three groups, ensuring that each question was annotated by two different annotators. As shown in Table~\ref{tab: human_label}, Cohen’s Kappa values exceeded 0.75 for confusion labels (where “Yes” indicates a question is out-of-scope and “No” indicates otherwise), and surpassed 0.81 for defusion labels (where “Yes” indicates the LLM${}_r$’s response defuses the question, and “No” indicates it does not), signifying substantial to near-perfect inter-annotator agreement and annotation consistency. Two additional annotators manually resolved all disagreements to establish ground truth labels. All annotators were computer science graduate students recruited via email and incentivized with complimentary meals. They were trained by first reviewing the annotation guidelines and then labeling a small dataset previously annotated by the authors. The annotation guidelines are released along with the data and code. 

We used two metrics to assess our method: “Annotator Acc” and “Group’s Agree Acc.” “Annotator Acc” treats each annotator’s labels as ground truth, computing the accuracy of the generated labels. “Group’s Agree Acc” uses the instances where both annotators within the same group agreed, treating this consensus as the ground truth. This method yielded higher accuracy than “Annotator Acc,” leading us to resolve disagreements with two additional annotators, ultimately establishing the final labeled data as ground truth.

As reflected in “Ground Truth Acc,” in Table ~\ref{tab: human_label}, our method achieved 94.91\% accuracy in generating out-of-scope and in-scope questions, while our proposed AutoDefusionEval method achieved 98.23\% accuracy in evaluating defusion. As shown in Figure~\ref{fig:confuse_defuse}, the confusion matrix on the left aligns closely with the ground truth, with minimal errors (5 false negatives and 6 false positives). The defusion matrix on the right reflects a similarly high accuracy, with only one error for false negatives and false positives. 

\begin{table}[H]  
\centering
\resizebox{0.5\textwidth}{!}{  
\begin{tabular}{c|cc}
\toprule
Model Name & Quantization & Knowledge Cutoff Date\\
\midrule
Llama 3.2 3B Instruct Turbo & FP8 & 2023/12 \\
Llama 3.1 8B Instruct Turbo & FP8 & 2023/12 \\
Llama 3.1 70B Instruct Turbo & FP8 & 2023/12 \\
Llama 3.3 70B Instruct Turbo & FP8 & 2023/12 \\
Mistral (7B) Instruct v0.3  & FP16 & 2024/5 \\
gpt-3.5-turbo & \hspace{6pt}--- & 2021/9 \\
\bottomrule
\end{tabular}
}
\caption{Different LLM${}_r$ being evaluated. In ELOQ-Silver, 66 news articles are published before Mistral 7B v0.3's knowledge cutoff date of 5/22/2024.}
\label{tab: diff_llm_r}
\end{table}

\section{Experimental Results}

\begin{table*}[t]  
\centering
\resizebox{\textwidth}{!}{  
\begin{tabular}{c|c|cccccccccc|cc}
\toprule

Prompt & Model &    business   &    entm  & food  & music & news & politics & science & sport & tech   & travel & Avg      & Std Dev\\
\midrule
\multirow{5}{*}{Basic} 
& GPT-3.5 &         18.06      &    10.96 & 12.32 & 14.27 & 13.59   &  15.95  & 10.15 & $18.37^\star$ & 15.59  &  \underline{9.63}  &  13.89   &  2.97\\
& Llama 3.2 3B &    63.59      &    61.55 & 61.09 & 62.07 & 63.49   &  64.01  & \underline{53.35} & $71.71^\star$ & 63.69  & 56.51  &  62.11   &  4.60\\
& Mistral 7B v0.3 & 68.40      &    67.63 & 60.88 & 67.74 & 67.96   &  66.86  & \underline{56.07} & $76.23^\star$ & 66.87  & 61.06  &  65.97   &  5.20\\
& Llama 3.1 8B &    71.25      &    69.42 & 67.56 & 72.04 & 67.16   &  \textbf{71.89}  & \underline{59.73} & $77.90^\star$ & 70.15  & 63.49  &  69.06   &  4.74\\
& Llama 3.1 70B &   \textbf{71.54}      &    70.22 & 67.56 & 71.07 & 67.66   &  70.75  & \underline{\textbf{61.51}} & $79.76^\star$ & 70.67  & 68.04  &  69.88   &  4.33\\
& Llama 3.3 70B &   71.25      &    \textbf{72.71} & \textbf{68.17} & \textbf{73.51} & \textbf{69.84}   &  69.90  & \underline{60.04} & $\textbf{82.42}^\star$ & \textbf{73.23}  & \textbf{68.78}  &  \textbf{70.98}   &  5.29\\
\midrule
\multirow{5}{*}{Two-shot}
& GPT-3.5 &         65.65      &    65.24 & 57.60 & 65.88 & 62.20   &  66.48  & \underline{52.41} & $71.71^\star$ & 59.90  & 56.83  &  62.39   &  5.43\\
& Llama 3.2 3B &    80.77      &    79.28 & 76.08 & 79.47 & 76.79   &  78.25  & 72.28 & $84.09^\star$ & 79.90  & \underline{71.64}  &  77.85   &  3.61\\
& Mistral 7B v0.3 & 80.86      &    77.69 & 74.54 & 78.79 & 79.17   &  79.11  & \underline{67.47} & $86.84^\star$ & 78.15  & 71.43  &  77.40   &  5.02\\
& Llama 3.1 8B &    83.32      &    80.98 & 80.18 & 84.16 & 78.77   &  \textbf{82.53}  & \underline{72.70} & $87.92^\star$ & 82.77  & 78.41  &  81.17   &  3.87\\
& Llama 3.1 70B &   85.18      &    \textbf{83.37} & \textbf{80.60} & \textbf{85.83} & \textbf{79.76}   &  82.24  & \underline{\textbf{75.84}} & $\textbf{90.57}^\star$ & 83.08  & \textbf{79.79}  &  \textbf{82.62}   &  3.84\\
& Llama 3.3 70B &   \textbf{85.48}      &    81.27 & 80.39 & 84.26 & 79.27   &  82.34  & \underline{\textbf{75.84}} & $88.90^\star$ & \textbf{83.49}  & 78.62  &  81.99   &  3.57\\
\midrule
\multirow{5}{*}{Zero-shot-CoT} 
& GPT-3.5 &         63.00      &    61.95 & 55.95 & 64.03 & 62.40   &  62.58  & \underline{53.03} & $71.71^\star$ & 59.69  & 56.30  &  61.07   &  4.96\\
& Llama 3.2 3B &    83.91      &    78.09 & 77.21 & 79.96 & 78.77   &  81.77  & \underline{73.85} & $85.46^\star$ & 80.41  & 75.77  &  79.52   &  3.39\\
& Mistral 7B v0.3 & 77.72      &    78.39 & 70.94 & 78.59 & 79.07   &  77.02  & \underline{67.26} & $84.77^\star$ & 76.41  & 72.06  &  76.22   &  4.69\\
& Llama 3.1 8B &    79.39      &    78.09 & 74.85 & 78.40 & 78.67   &  79.20  & \underline{70.29} & $83.89^\star$ & 77.64  & 74.50  &  77.49   &  3.43\\
& Llama 3.1 70B &   90.28      &    87.85 & 87.27 & 88.56 & 86.51   &  89.08  & \underline{83.68} & $91.55^\star$ & 86.97  & 85.19  &  87.69   &  2.21\\
& Llama 3.3 70B &   \textbf{93.52}      &    \textbf{91.24} & \textbf{90.55} & \textbf{92.57} & \textbf{90.77}   &  \textbf{92.12}  & \underline{\textbf{89.12}} & $\textbf{95.09}^\star$ & \textbf{93.85}  & \textbf{90.90}  &  \textbf{91.97}   &  1.71\\
\bottomrule
\end{tabular}
}
\caption{Evaluation of LLMs' accuracy in defusing out-of-scope questions from the ELOQ-Silver dataset across diverse news topics. For each LLM, the \underline{underscored} value and starred ($^\star$) value are the minimum and maximum values, respectively, across all the topics within the same prompting method. \textbf{Bold} values are the maximum values for each topic across all the LLMs within the same prompting method. ``entm'' is the abbreviation of ``entertainment''.}
\label{tab: confusion_response}
\end{table*}

\subsection{Out-of-Scope Response} \label{sec: exp_conf_response}

In this section, we examine LLMs’ ability to defuse out-of-scope questions. The selected LLMs are listed in Table~\ref{tab: diff_llm_r}. Most of these models have a knowledge cutoff date of December 2023, which predates the publication of the news articles in ELOQ. As a result, they must generate responses based solely on their understanding of the provided news content. As shown in Table~\ref{tab: confusion_response}, we evaluated LLMs on their ability to defuse out-of-scope questions with three prompting methods (Appendix~\ref{ap: prompt_rag}, ~\ref{ap: prompt_rag_twoshot}, ~\ref{ap: prompt_rag_cot}). As shown in Table \ref{tab: confusion_response}, a) Two-shot prompt boosts all evaluated LLMs' accuracy substantially, especially GPT-3.5, likely due to the examples clarifying how to respond to out-of-scope questions. In contrast, Zero-shot-CoT \cite{DBLP:conf/nips/KojimaGRMI22} merely invites the LLM to ``reason step by step’’ beats Two-shot on three out of six LLMs.
b) Larger models (70B) benefit more from Zero-shot-CoT than from Two-shot prompting, achieving accuracy gains of 5.07\% to 9.98\%, whereas smaller models (3B to 8B) show only marginal changes, ranging from -3.68\% to 1.67\%. This difference likely stems from larger models’ ability to better utilize their reasoning capabilities and extensive knowledge;
c) Most LLMs perform worst on the ``science'' topic and best on ``sport’’. This is expected, as scientific content tends to contain more implicit premises assumed by the author rather than explicitly stated facts, thus requiring more domain knowledge. In contrast, sports content is simpler and unambiguous, with facts often clearly stated in the document.

\subsection{Out-of-Scope Detection} \label{sec: exp_conf_detection}

\begin{figure}[t]
    \centering
    \includegraphics[width=\linewidth]{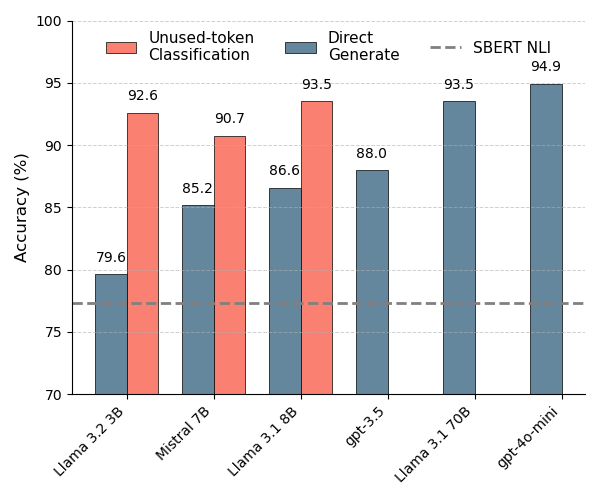}
    \caption{Evaluation of out-of-scope detection on ELOQ-Gold.}
    \label{fig: diff_llm_confuion}
\end{figure}

The primary challenge in preventing an LLM from hallucinating or generating incorrect answers to out-of-scope questions is detecting such scenarios effectively. Once we can accurately identify these cases, we can handle them easily by using a prompt and in-context examples specifically designed for out-of-scope questions. We use two methods for out-of-scope detection. First is a prompt-based method where we simply prompt the LLM to do such detection (Appendix~\ref{ap: prompt_confusion}), denoted as ``Direct Generate’’ in Figure \ref{fig: diff_llm_confuion}. Second is a classification method where we train a binary classifier on ELOQ-Silver using the embeddings of an appended unused-token as features, denoted as ``Unused-token Classification’’ in Figure \ref{fig: diff_llm_confuion}. 

\subsubsection{Unused-token Classification} 
The out-of-scope detection problem can be framed simply as a binary classification task, where we determine if a given question $q$ is out-of-scope or not for a given document $d$. The binary label is $y \in \{0,1\}$ where $y=1$ indicates the question is relevant to the document and $y=0$ indicates the question is out-of-scope. The objective is to learn a classification function: $f:(q,d) \mapsto p(y=1 \mid q,d)$, which estimates the likelihood of the question-document pair to be in-scope. We would like to test whether the LLM's internal features contain enough information to classify a question-document pair effectively to simulate the effect that the LLM ``realizes'' the out-of-scope scenario before generating a response. To achieve this, we use the sequence representation of the entire input -- denoted as $x=\text{concat}(I,q,d,)$ where $I$ is the instruction prompt provided to the LLM (e.g. ``Can this document answer the question?''). We extract the sequence representation by appending a reserved unused-token to the end of the input sequence and extracting its representation from the final hidden state of the last transformer layer. The rationale behind using unused-token is that, since LLM developers reserve these tokens for potential future use in specialized training tasks, they are untrained and uninfluenced by any prior data. Thus, the unused-token does not have any predefined meanings and can be viewed as a blank buffer, aggregating interactions among all the input tokens and capturing an uninfluenced representation of the sequence. We also tried using the \texttt{<eos>} token, which yielded similar results. 

The classifier is a two-layer MLP that takes the unused-token representation as input and is optimized against the binary label $y$. This approach is commonly known as ``probing'' \cite{DBLP:journals/corr/abs-2407-20311}, where the LLM weights are frozen, and the classifier can be viewed as an external probe to assess the information contained in the model's internal representations. The performance reported in Figure \ref{fig: diff_llm_confuion} was tested on ELOQ-Gold, the human-labeled data. For more details about the training, please refer to Appendix~\ref{ap: embedding_classifier}. We also provide a ``Sentence BERT NLI'' baseline where we adopt NLI style training using the off-the-shelf \footnote{\url{https://huggingface.co/sentence-transformers/all-mpnet-base-v2}} model from HuggingFace. Specifically, we follow \cite{DBLP:conf/emnlp/ReimersG19} to extract document and question \texttt{[CLS]} embeddings $e_d$ and $e_q$ to use $[e_d; e_q; |e_d - e_q|]$ as the input features. 

\subsubsection{Results Discussion}
For the ``Direct Generation'' approach, we can observe a clear positive correlation between model size and accuracy, aligning with the general trend of increased capability with larger models. However, due to the proprietary nature of OpenAI's models and computational constraints, we are only able to obtain hidden states to train the binary classifier for three LLMs: Llama 3.2 3B, Mistral 7B, and Llama 3.1 8B. Notably, the classifiers consistently outperform the ``Direct Generation'' method, indicating that the LLM's internal representations encode the necessary information for out-of-scope detection, yet the model may fail when directly prompted. Our dataset, ELOQ, demonstrates its usefulness by enabling even smaller models -- such as Llama 3.2 3B-based classifier -- to achieve accuracy comparable to significantly larger models that rely on direct generation, including Llama 3.1 70B and GPT-4o-mini. Moreover, as shown in Figure \ref{fig: diff_llm_confuion}, all models and methods outperform the ``SBERT NLI'' baseline, suggesting that out-of-scope detection is non-trivial and requires a deeper and more complex understanding of linguistic relationships that larger models capture during pretraining and fine-tuning. Lastly, in our exploratory experiments, we trained a classifier on GPT-2, but its accuracy is no better than random guessing, further highlighting the importance of model capacity in leveraging internal representations for classification. 

\begin{table}[h]
    \centering
    \begin{tabular}{lcccc}
        \toprule
        \textbf{Model} & \textbf{Recall@1} & \textbf{Recall@5} & \textbf{Recall@10} & \textbf{MRR} \\
        \midrule
        BM25 \cite{DBLP:journals/corr/abs-0911-5046}  & 0.4637 & 0.6880 & 0.7487 & 0.5653 \\
        SBERT \cite{DBLP:conf/emnlp/ReimersG19}  & 0.4685 & 0.7058 & 0.7858 & 0.5773 \\
        BGE \cite{bge_embedding}     & 0.5326 & 0.7604 & 0.8267 & 0.6352 \\
        Stella \cite{DBLP:journals/corr/abs-2412-19048}  & 0.4875 & 0.7376 & 0.8128 & 0.5994 \\
        Linq \cite{LinqAIResearch2024}  & 0.5603 & 0.7890 & 0.8492 & 0.6618 \\
        \bottomrule
    \end{tabular}
    \caption{Retrieval performance of out-of-scope questions on ELOQ}
    \label{tab: semantic_relevance}
\end{table}

\subsection{Semantic Relevance}

To demonstrate that the generated out-of-scope questions are indeed semantically similar to the document from which they are generated, we conduct retrieval experiments using all out-of-scope questions against our document corpus (2,000 documents). We employ the following retrieval models: BM25 \footnote{\url{https://github.com/castorini/pyserini}}, SBERT \footnote{\url{https://huggingface.co/sentence-transformers/all-mpnet-base-v2}}, \texttt{bge\allowbreak-large\allowbreak-en\allowbreak-v1.5} \footnote{\url{https://huggingface.co/BAAI/bge-large-en-v1.5}}, \texttt{stella\allowbreak\_en\allowbreak\_1.5B\allowbreak\_v5}\footnote{\url{https://huggingface.co/NovaSearch/stella_en_1.5B_v5}}, and \texttt{Linq\allowbreak-Embed\allowbreak-Mistral}\footnote{\url{https://huggingface.co/Linq-AI-Research/Linq-Embed-Mistral}}. Among these models, BM25 and SBERT are well-known retrievers that serve as established baselines, \texttt{bge\allowbreak-large\allowbreak-en\allowbreak-v1.5} was accepted in SIGIR 2024 resource track as a high-performance embedding model, \texttt{stella\allowbreak\_en\allowbreak\_1.5B\allowbreak\_v5} ranks \#1 on MTEB's English v1 dataset \cite{muennighoff2022mteb}, and Linq-Embed-Mistral leads multiple benchmarks, including BEIR \cite{kamalloo2023resources} and MTEB's English v2 dataset at the time of writing. As shown in Table \ref{tab: semantic_relevance}, the Recall@1 values across all models suggest that for approximately half of the out-of-scope questions, the top-retrieved document is the same one from which the question was generated. When we relax this condition to Recall@10, around 80\% of out-of-scope questions retrieve their associated document within the top 10 results. This demonstrates a strong semantic connection between out-of-scope questions and their source documents, despite the fact that the document lacks the necessary information to answer the question.

\begin{table}[t!]  
\centering
\small
\begin{tabular}{l|c}
\toprule
Module & Accuracy\\
\midrule
Base (gpt-4o-mini) & 87.61\\
+ examples & 97.35\\
+ self-consistency (m=3) & 98.23\\
+ self-consistency (m=9) & 98.23\\
\bottomrule
\end{tabular}
\caption{Evaluation of different prompts for ``defusion detection'' on ELOQ-Gold.}
\label{tab: ab}
\end{table}
\subsection{Ablation Study} \label{sec: ab_study}

We examined the impact of various prompts on ``defusion detection” task. As seen in Table \ref{tab: ab}, incorporating examples increased accuracy from 87.61\% to 97.35\%, and applying self-consistency with $m=3$ further boosted performance to 98.23\%. However, further increasing $m$ did not lead to additional improvements. Since ELOQ-Gold is a smaller dataset, using $m=9$ did not yield gains, but this doesn’t rule out its potential benefit on the larger ELOQ-Silver dataset. Therefore, we adopted $m=9$ in our experiments to balance accuracy and expenses (Appendix~\ref{ap: expense}).

\section{Conclusion and Future Work}
We introduced ELOQ, a benchmark for enhancing LLM detection of out-of-scope questions for a given document. Human annotation validates that ELOQ achieves 94.91\% accuracy in generating out-of-scope and in-scope questions, while our AutoDefusionEval achieves 98.23\% accuracy in detecting effective defusion responses. We further demonstrated ELOQ’s utility by training out-of-scope detectors, and the results show that a small model (Llama 3.1 8B) can achieve comparable results to the larger one (Llama 3.1 70B) without training. For future work, we plan to apply for OpenAI funding to utilize more powerful models, such as o1, to improve both data quality and scale. Additionally, we aim to explore post-training techniques to fine-tune LLMs on ELOQ, ensuring they align better with human preferences by identifying the confusing aspects of a question rather than generating hallucinated answers.

\begin{acks}
We would like to express our sincere gratitude to Xiaoxiao Shang, Ethan Lin, Wei Mo, Zhan Shi, Xinze Ye, and Haisong Wang for their invaluable contributions to data labeling.
\end{acks}

\section*{Appendix}
\appendix

\section{Expenses}\label{ap: expense}
Generating the ELOQ dataset using GPT-4o-mini costs approximately \$80. For LLMs not provided by OpenAI, we used the API service from \url{https://www.together.ai/}. Excluding the dataset generation, all other experiments cost around \$100.

\section{Prompt}\label{ap: prompt}
\subsection{Extract Claims}\label{ap: extract_prompt}

\noindent You will be provided with a document delimited by triple quotes. Read the document and follow user's instructions.\\  
Read the document and list \{num\_fact\} most important facts it contains. Each fact should be stated in a clear, standalone sentence with sufficient context to be understood independently, avoiding undefined pronouns. Ensure that each fact is directly derived from the document and does not include any information not mentioned within it.\\  

\noindent Document:\\ 
"""\{document\}""" \\

\noindent \{num\_fact\} most important facts:

\subsection{Recover Missing Claims}\label{ap: recover_prompt}
\noindent Read the document below with a list of \{num\_fact\} facts it contains. Note that some of the facts are missing represented by (missing). Your task is to guess the missing facts could have said and complete the missing facts. Each fact should be stated in a clear, standalone sentence with sufficient context to be understood independently, avoiding undefined pronouns.
Please complete the missing facts and return all the \{num\_fact\} facts in original order. You must return all the \{num\_fact\} facts.\\\\
Document:\\
"""{document}"""\\\\
Completed list of facts:

\subsection{Out-of-scope Judgement}\label{ap: prompt_confusion}
\noindent You will be provided with a document delimited by triple quotes.
Read the document and follow user's instructions.\\
Read the document and the question. Check if the question contains any confusing parts: mentions of people, locations, organizations, or other entities that are not present in the document\\\\
Document:\\
"""\{document\}"""\\\\
Question:\\
\{question\}\\\\
Does the question contain any confusing parts: mentions of people, locations, organizations, or other entities that are not present in the document? Please start by explaining your reasoning. End your response with 'The answer is: Yes or No.':

\subsection{Response Defusion Judgement}\label{ap: prompt_defusion}
\noindent You will be provided with a document delimited by triple quotes.
Read the document and follow user's instructions.
Read the document and answer the question based on the document.\\
Document:\\
"""\{example\_document\}"""\\\\
Question:\\
\{example\_questions\}\\\\
Answer:\\
\{example\_LLM\_responses\}\\\\
\{Defusion Instruction (Appendix ~\ref{ap: defusion_instruction})\}\\
\{example\_defuse\_reasoning\}\\\\
Read the document and answer the question based on the document.\\\\
Document:\\
"""\{document\}"""\\\\
Question:\\
\{question\}\\\\
Answer:\\
\{LLM\_response\}\\\\
\{Defusion Instruction (Appendix ~\ref{ap: defusion_instruction})\}

\subsection{Defusion Instruction}\label{ap: defusion_instruction}
\noindent The question presented is confusing and unanswerable based on the document provided. Read the response carefully. If the response attempts to answer the question, like answering 'Yes' to agree with the question or 'No' to disagree with the question, you should immediately conclude with ‘The answer is: No.’ even if it later acknowledges the question’s confusion. Otherwise, determine if the response clarifies that the question cannot be answered or if the response indicates that the information needed to answer the question is not available or not mentioned in the document. Begin with your reasoning and finally conclude with ‘The answer is: Yes.’ or ‘The answer is: No.’. Note that if the response attempts to answer the question at any point, like answering 'Yes' to agree with the question or 'No' to disagree with the question, you should immediately conclude with ‘The answer is: No.’ even if it later acknowledges the question’s confusion:

\subsection{RAG Basic}\label{ap: prompt_rag}
\noindent You will be provided with a document delimited by triple quotes.
Read the document and follow user's instructions.\\
Read the document and answer the question based on the document.\\\\
Document:\\
"""\{document\}"""\\\\
Question:\\
\{question\}\\\\
Answer: 

\subsection{RAG Two-shot}\label{ap: prompt_rag_twoshot}
\noindent You will be provided with a document delimited by triple quotes.
Read the document and follow user's instructions.\\
Read the document and answer the question based on the document.\\\\
Document:\\
"""\{example\_document\}"""\\\\
Question:\\
\{two\_example\_questions\}\\\\
Answer:\\
\{two\_example\_responses\}\\\\
Read the document and answer the question based on the document.\\\\
Document:\\
"""\{document\}"""\\\\
Question:\\
\{question\}\\\\
Answer: 

\subsection{RAG Zero-shot-CoT}\label{ap: prompt_rag_cot}
\noindent You will be provided with a document delimited by triple quotes.
Read the document and follow user's instructions.\\
Read the document and reason step by step to answer the question based on the document. If the question cannot be answered using the document, state explicitly that the question cannot be answered.\\\\
Document:\\
"""\{document\}"""\\\\
Question:\\
\{question\}\\\\
Answer: 

\subsection{In-scope Question Generation}\label{ap: prompt_non_conf_q_gen}
\noindent You will be provided with a document delimited by triple quotes.
Read the document and follow user's instructions.
Read the document attentively and compile a numbered list of the top \{num\_q\} questions that the document directly answers. Ensure each question is clear, accurate, and devoid of confusion, false assumptions, undefined pronouns, or misinformation. Avoid referencing people, locations, organizations, or other entities not explicitly mentioned in the document. Construct each question to be thought-provoking, containing between 13 to 18 words, and sufficiently detailed to avoid being overly straightforward.\\\\
Document:\\
"""\{document\}"""\\\\
Questions:

\subsection{Out-of-scope Question Generation}\label{ap: prompt_conf_q_gen}
\noindent You will be provided with a document delimited by triple quotes.
Read the document and follow user's instructions.\\
Read the document and review the list of hallucinated facts. For each hallucinated fact, craft a single, specific and concise question containing 13 to 18 words that incorporate the key element of the fact, ensuring the question is intentionally confusing. The question should not be answerable using any information present in the document. The question should not combine multiple queries and each question should address only 
one specific aspect. If a question cannot be formulated for a particular hallucinated fact, you may omit it.\\\\
Document:\\
"""\{document\}"""\\\\
hallucinated facts:\\
\{hallucinated\_facts\}\\\\
Questions:

\subsection{Remove Claims}\label{ap: prompt_remove_claim}
\noindent You will be provided with a document delimited by triple quotes.
Read the document and follow user's instructions.\\
Read the document below with a list of \{num\_true\_fact\} ground-truth facts it contains and a list of \{num\_false\_fact\} hallucinated facts that are not supported by the document. Your task is to remove any hallucinated facts that can be supported by either the document or the \{num\_true\_fact\} ground-truth facts. Please only return the remaining hallucinated facts, along with their original order numbers.\\\\
Document:\\
"""\{example\_document\}"""\\\\
\{num\_true\_fact\} ground-truth facts:\\
\{true\_facts\}\\\\
\{num\_false\_fact\} hallucinated facts:\\
\{false\_facts\}\\\\
Remaining hallucinated facts:\\
\{remained\_facts\}\\\\
Read the document below with a list of \{num\_true\_fact\} ground-truth facts it contains and a list of \{num\_false\_fact\} hallucinated facts that are not supported by the document. Your task is to remove any hallucinated facts that can be supported by either the document or the \{num\_true\_fact\} ground-truth facts. Please only return the remaining hallucinated facts, along with their original order numbers.\\\\
Document:\\
"""\{document\}"""\\\\
\{num\_true\_fact\} ground-truth facts:\\
\{true\_facts\}\\\\
\{num\_false\_fact\} hallucinated facts:\\
\{hallucinated\_facts\}\\\\
Remaining hallucinated facts:

\section{Embedding Classifier Implementation Details}\label{ap: embedding_classifier}
We train the classifier on ELOQ-Silver, which is split as: 80\% for training, 10\% for evaluation, and 10\% for testing. After training, we test the classifier on ELOQ-Gold data. 

For the Llama family, we use \texttt{<|reserved\allowbreak\_special\allowbreak\_token\allowbreak\_0|>} as the unused-token. This token is not trained and is reserved for future use cases such that the users do not need to resize the vocabulary. For Mistral-7B-v0.3, we use \texttt{[control\_555]} as the unused-token because Mistral's official documentation indicated that they reserve 768 control tokens for future use, and none of them are used during training. 555 is a random choice. 

The unused tokens ' hidden state is used as input features for a 2-layer MLP, which is optimized with a binary cross-entropy (BCE) loss. A dropout of 0.1 is applied after the first layer. We use Adam optimizer with 1e-4 learning rate. We train each classifier for 10 epochs with batch size 8. The best validation checkpoint is saved and its performance is reported on ELOQ-Gold. 

\section{Example News}\label{ap: exp_oos_doc}

Justin Jefferson, CeeDee Lamb, NFL Injury Statuses and Fantasy Impact for Week 3 Stephen Maturen/Getty Images

Two of the best wide receivers in the NFL are expected to be on the field for their respective Week 3 games. Justin Jefferson and CeeDee Lamb each received positive prognosis about their injury issues, which is something that can't be said for the rest of the stars across the NFL. Christian McCaffrey is already on injured reserve, Deebo Samuel is out for a weeks and a slew of other running backs and wide receivers are dealing with ailments that could keep them out of Week 3. Below is a look at all of the significant injuries that could affect fantasy football matchups across Week 3. Justin Jefferson Off Injury Report

Justin Jefferson was taken off the Minnesota Vikings injury report on Friday. Jefferson's status was up in the air because of a quad injury, but he practiced well enough this week that the injury is not a concern. The superstar wide out will be needed for Minnesota's home clash with the Houston Texans, which has the potential to be a high-scoring affair. Jefferson is always the primary target in Minnesota when healthy, but he should have more targets in Week 3 because Jordan Addison and T.J. Hockenson are still out. Jefferson earned seven targets from Sam Darnold in Week 2. Only running back Aaron Jones had more than four targets against the New York Giants. Houston's defense may give Jefferson some fits, led by cornerback Derek Stingley Jr. but it has allowed 330 receiving yards to opposing wide outs on just 20 catches through two weeks. Jefferson is an automatic start whenever he's healthy, and his star power may be needed more on certain fantasy rosters in Week 3 depending on how many injuries affect a single roster.
\clearpage
\newpage
\balance
\bibliographystyle{ACM-Reference-Format}
\bibliography{sample-base}

\end{document}